\documentclass{article}
\usepackage{color}
\usepackage[margin=1in]{geometry}
\usepackage{graphicx}
\usepackage{amssymb} 
\usepackage{amsmath} 
\usepackage{hyperref}
\usepackage[numbers]{natbib}

\title{MLE-induced Likelihood for Markov Random Fields}

\author{Jie Liu$^1$\footnote{Corresponding author's email address is jieliu@cs.wisc.edu}, Hao Zheng$^2$\\
\\
$^1$ Department of Computer Sciences \\
$^2$ Department of Statistics \\
University of Wisconsin, Madison}

\begin{document}

\maketitle

\begin{abstract}
Due to the intractable partition function, the exact likelihood function for a Markov random field (MRF), in many situations, can only be approximated.
Major approximation approaches include pseudolikelihood \citep{besag75} and Laplace approximation \citep{welling06}.
In this paper, we propose a novel way of approximating the likelihood function through first approximating the marginal likelihood functions of individual parameters and then reconstructing the joint likelihood function from these marginal likelihood functions.
For approximating the marginal likelihood functions, we derive a particular likelihood function from a modified scenario of coin tossing which is useful for capturing how one parameter interacts with the remaining parameters in the likelihood function.
For reconstructing the joint likelihood function, we use an appropriate copula to link up these marginal likelihood functions.
Numerical investigation suggests the superior performance of our approach.
Especially as the size of the MRF increases, both the numerical performance and the computational cost of our approach remain consistently satisfactory, whereas Laplace approximation deteriorates and pseudolikelihood becomes computationally unbearable.
\end{abstract}

\section{Introduction}

Suppose that we observe $n$ i.i.d.~data points $\mathbb{X}{=}\{ {\bf x}^1,...,{\bf x}^n\}$ sampled from a Markov random field (MRF) on ${\bf X} {\in} \mathcal{X}^d$ ($\mathcal{X}$ is a discrete space) which is parameterized by ${\boldsymbol \theta}$. 
The likelihood function $L({\boldsymbol \theta}|\mathbb{X})$ plays a key role in statistical inference of an MRF, including parameter estimation, model comparison, simulation practice and others. 
However due to the intractable partition function, the exact likelihood function, in many situations, can only be approximated. 
Widely used approximation approaches include pseudolikelihood \citep{besag75}, Laplace approximation \citep{welling06} and their variations, but these methods may suffer from either precision loss or a heavy computational burden, especially when we are dealing with large-scale MRFs.

Aiming at a fast and reliable approximation of the MRF likelihood function, we propose a two-step method as follows. 
First, we focus on the individual dimensions of the parameter space, and approximate the marginal likelihood \citep{mackay03} of the individual parameters. 
We derive a particular likelihood function from a modified scenario of coin tossing, and use this function for marginal likelihood approximation. 
In the second step, we reconstruct the joint likelihood function by linking up these marginal likelihood functions. 
We refer to a popular technique in high-dimensional statistical inference --- copula --- to reconstruct the joint likelihood. 
Because the first step of our approximation depends on the availability of the maximum likelihood estimate (MLE) of the parameters, we term the achieved likelihood function {\it MLE-induced likelihood}.

The rationale of our two-step approximation of MRF likelihood is as follows. 
In the first step of approximation, we only focus on the marginal likelihood functions and make sure that they are accurately approximated.
The particular likelihood function (derived from the modified scenario of coin tossing) properly captures how one parameter interacts with the remaining parameters (as a whole) in the likelihood function.
This is more effective than modeling how one parameter interacts with the remaining parameters individually.
In addition to this particular function, the availability of MLE for parameters as well as the validness of mode mapping makes sure that the marginal likelihood functions are accurate.
More important, the second step of reconstruction is supported by Sklar's Theorem which states that any multivariate joint distribution can be written in terms of univariate marginal distribution functions and a copula which describes the dependence structure between the variables \citep{sklar59}. 
Therefore, the second step of reconstruction preserves the precision of the individual marginal likelihood functions, which guarantees that the reconstructed joint likelihood is marginal-wise accurate. 
Third, the two-step approximation is also computationally desirable. 
The major computation cost is from approximating the individual marginal likelihood functions, which is linear in the dimension of the parameter space, making it particularly attractive for large-scale MRFs. 

The rest of the paper is organized as follows. 
Section \ref{sec:mle-inducedL} systematically presents how we derive this particular likelihood function for marginal likelihood approximation, and how we reconstruct the joint likelihood function via copula.
Section \ref{sec:simulation} empirically demonstrates that our MLE-induced likelihood works both more accurate and faster than pseudolikelihood \citep{besag75} and Laplace approximation \citep{welling06}.
Especially as the size of MRFs increases, both the numerical performance and the computational cost of our approach remain consistently satisfactory, whereas the approximation from Laplace approximation deteriorates and pseudolikelihood becomes computationally unbearable. 
We finally conclude in Section \ref{sec:conclusion}.

\section{MLE-induced Likelihood}\label{sec:mle-inducedL}

In the proposed MLE-induced likelihood, we first approximate the marginal likelihood functions of the individual parameters, and then reconstruct the joint likelihood function using a copula function \citep{nelsen06}. 
Subsection \ref{sec:generalizedBeta} introduces a particular parametric likelihood function derived from a modified scenario of coin tossing. 
Subsection \ref{sec:approximateMarginalLikelihood} explains how we approximate the marginal likelihood functions with the particular likelihood function, given the availability of the MLE of the parameters. 
Subsection \ref{sec:copula} completes our approximation by linking up the individual marginal likelihood functions via an appropriate copula.

\subsection{A Modified Scenario of Coin Tossing}\label{sec:generalizedBeta}

Suppose that we have a coin with a probability $\theta$ $(0{<}\theta{<}1)$ of landing ``head'' in a random toss. Let $X^i \in \{0,1\}, 1 \leq i \leq n$, denote the landing results, ``tails'' or ``heads'', from $n$ independent tosses of the coin. 
$X^i$ follows a Bernoulli distribution, and the total number of ``heads" $\sum_{i=1}^{n} X^i$ follows a Binomial distribution. 
We slightly change the conventional setting by casting an external effect, say, a magnetic field, to the coin tossing, so that the probability of landing ``head" in a random toss becomes $\lambda$, as specified by

\begin{equation}\label{eq:multiplicative}
\lambda = \frac{\eta_0\cdot\theta}{\eta_0\cdot\theta+(1-\eta_0)(1-\theta)},
\end{equation}

\noindent
where $\eta_0$ $(0<\eta_0<1)$ measures the external effect. 
$\eta_0$ is multiplicative to $\theta$ in the sense given in (\ref{eq:multiplicative}). 
When $\eta_0$ is $0.5$, this modified scenario reduces to the conventional coin tossing problem.

Under the external effect $\eta_0$, suppose that we observe the outcomes from the $n$ random tosses, denoted by $\mathbb{X}{=}\{ {x}^1,...,{x}^n\}$. 
The likelihood function of $\lambda$ is

\begin{equation}\label{eq:likelihoodOfLambda}
L(\lambda|\mathbb{X}) = P(\mathbb{X} | \lambda) = \binom{n}{\alpha_0} \; \lambda^{\alpha_0} \; (1-\lambda)^{n-\alpha_0},
\end{equation}

\noindent
where $\alpha_0 = {\sum_{i=1}^n x^i}$. Plugging (\ref{eq:multiplicative}) into (\ref{eq:likelihoodOfLambda}), we have the likelihood function of $\theta$

\begin{equation}\label{eq:specialbeta}
\begin{split}
L(\theta|\mathbb{X}) = \binom{n}{\alpha_0} \; & \eta_0^{\alpha_0}(1-\eta_0)^{n-\alpha_0} \\
 & \; \frac{ \theta^{\alpha_0}(1-\theta)^{n-\alpha_0} }
{(\eta_0 \cdot\theta + (1-\eta_0)(1-\theta))^n} \; .
\end{split}
\end{equation}

This particular likelihood function (\ref{eq:specialbeta}) is very useful in approximating the marginal likelihood functions of an MRF, as explained next. 

\subsection{Approximating Marginal Likelihood of an MRF}\label{sec:approximateMarginalLikelihood}

\begin{figure}[h]
\begin{center}
(a)\includegraphics[trim=0cm 2.6cm 0cm 1cm, clip=true, angle=0, scale=0.28]{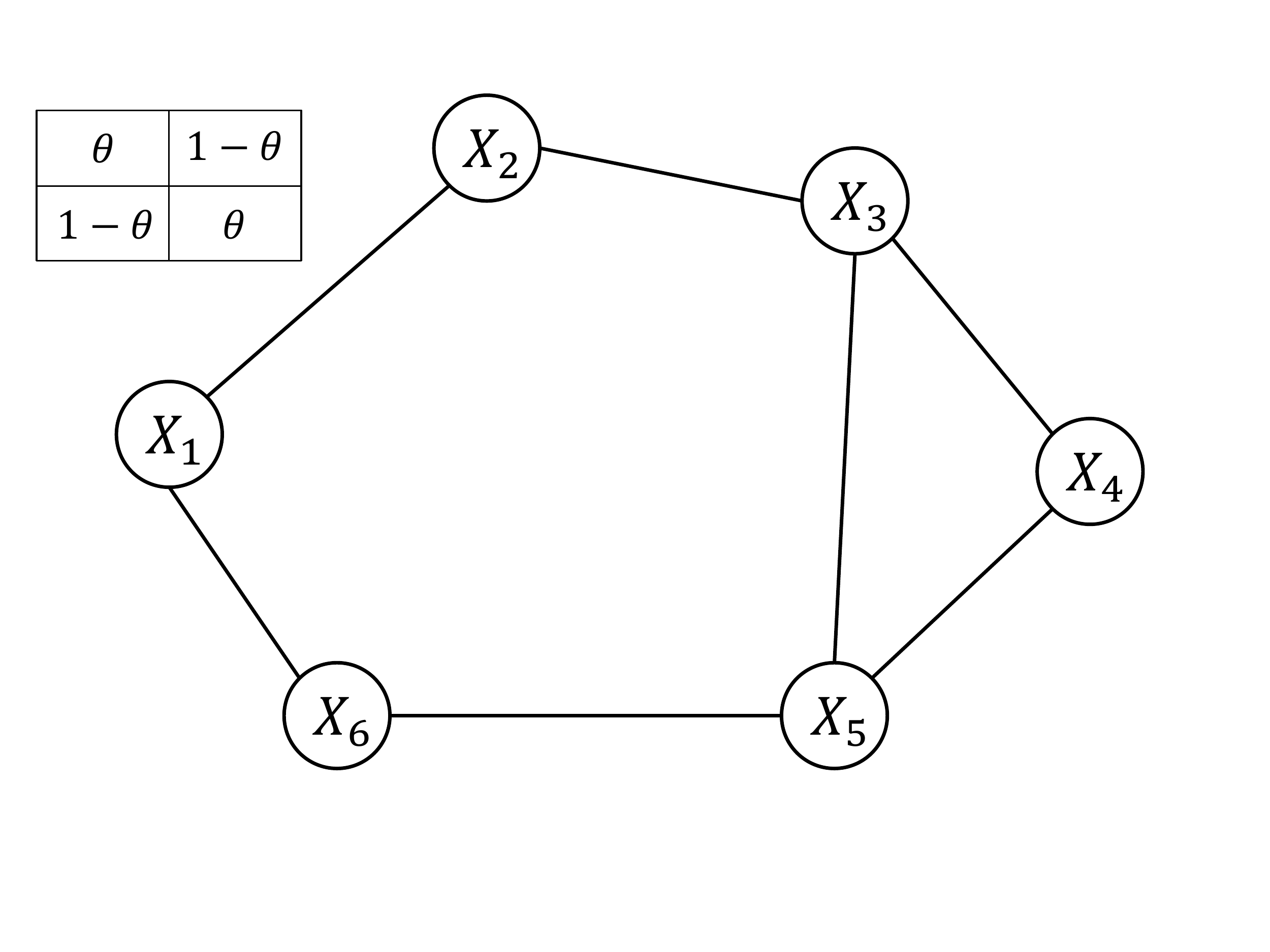} \qquad
(b)\includegraphics[trim=0cm 2.6cm 0cm 1cm, clip=true, angle=0, scale=0.28]{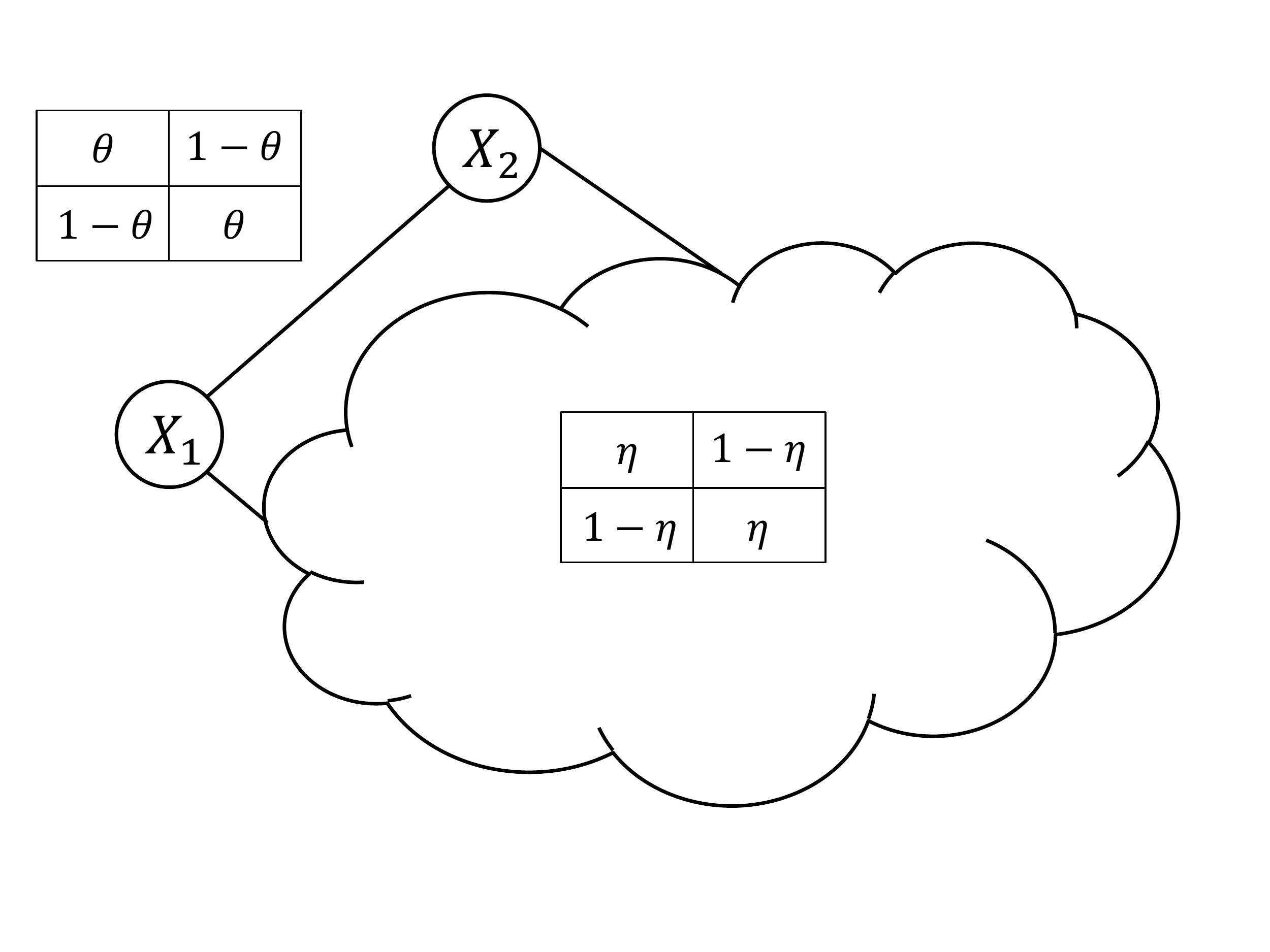}
\end{center}
\caption{(a) An illustrative pairwise MRF $\mathcal{M}$. (b) Another representation of $\mathcal{M}$ by masking the part of $\mathcal{M}$ other than $\{X_1,X_2\}$. The two pairwise potential functions quantify the probabilities that $X_1$ and $X_2$ agree with each other.
}
\label{fig:graphicalModel}
\end{figure}

This subsection shows that it is natural to approximate the marginal likelihood functions of an MRF with the particular likelihood function given in (\ref{eq:specialbeta}). 
For illustration, we use a binary pairwise MRF $\mathcal{M}$ with $d$ nodes and $p$ parameters. 
The pairwise potential function $\phi_j$ on edge $j$ (connecting $X_u$ and $X_v$) parameterized by $\theta_j$ ($0{<}\theta_j{<}1$) is $\phi_j({\bf X};\theta_j)=\theta_j^{I_{\{X_u=X_v\}}}(1-\theta_j)^{I_{\{X_u \not=X_v\}}}$. 
The exact structure of $\mathcal{M}$ is provided in Figure \ref{fig:graphicalModel}(a).
$\mathcal{M}$ includes six nodes and seven edges. 

As shown in Figure \ref{fig:graphicalModel}(a), $\theta$ parameterizes the edge that directly connects $X_1$ and $X_2$. 
The corresponding pairwise potential function is $\phi({\bf X};\theta)=\theta^{I_{\{X_1=X_2\}}}(1-\theta)^{I_{\{X_1\not=X_2\}}}$. 
Besides this edge, $X_1$ and $X_2$ influence each other through the remainder of the network. 
Even for such a small MRF, it is difficult to derive the impact from the remaining part of $\mathcal{M}$ towards the concurrence of $X_1$ and $X_2$. 
To circumvent the difficulty of explicitly quantifying the overall effect to $\{X_1,X_2\}$, we simplify the impact from the remainder of the network regarding the $X_1, X_2$ concurrence as follows. 

As illustrated in Figure \ref{fig:graphicalModel}(b), we mask the remaining part outside $\{X_1,X_2\}$.
We assume that the effect from the remaining network can be captured by another potential function. 
Then $\mathcal{M}$ is simplified as two nodes $X_1$ and $X_2$ and two edges connecting them, one direct edge and one indirect edge masked in the cloud. 
Similar to the parameterization of the original direct edge with $\theta$, we introduce another parameter $\eta$ to parameterize the potential function on this imaginary indirect edge, which corresponds to the remainder of $\mathcal{M}$. 
From the probability density function of the MRF (a.k.a. a log linear model), $\theta$ and $\eta$ are multiplicative to each other, and the combined effect can be parameterized by  

\begin{equation}\label{eq:multiplicativeRandom}
\lambda = \frac{\eta\cdot\theta}{\eta\cdot\theta+(1-\eta)(1-\theta)}.
\end{equation}

Given the $n$ observed data points $\mathbb{X}$ from $\mathcal{M}$, the marginal likelihood function of $\theta$ can be derived by marginalizing out all of the other parameters. 
By simplifying the remainder of the network into the indirect edge, we lump all of the other parameters into one single parameter $\eta$. 
Because it is not feasible in general to obtain the accurate distribution of $\eta$, we propose using its point estimate $\eta_0$ to approximately quantify its distribution. 
Therefore, we can regard the effect from the remaining network $\eta_0$ as the external effect in the modified scenario of coin tossing.
Following the derivation in Subsection \ref{sec:generalizedBeta}, the marginal likelihood function of $\theta$ has a form of (\ref{eq:specialbeta}). 
If the point estimate $\eta_0$ is provided, the approximated marginal likelihood function can be easily calculated.

The remaining issue is to find the point estimate $\eta_0$.
We rely on the availability of MLE and the fact that ``MLE is invariant to reparameterization" {\citep{shaomathematical}}. 
If the MLEs for $\theta$ and $\lambda$ (denoted by $\hat{\theta}_{\text{MLE}}$ and $\hat{\lambda}_{\text{MLE}}$) are available, we are able to calculate $\eta_0$ by rearranging (\ref{eq:multiplicativeRandom}) as

\begin{equation} \label{eq4}
{\eta}_0 = \frac{\hat{\lambda}_{\text{MLE}} \cdot (1-\hat{\theta}_{\text{MLE}})} {\hat{\lambda}_{\text{MLE}} + \hat{\theta}_{\text{MLE}} - 2 \: \hat{\lambda}_{\text{MLE}} \cdot \hat{\theta}_{\text{MLE}}}.
\end{equation}

In terms of calculation, it is straightforward to show that

\begin{equation} \label{eq5}
\hat{\lambda}_{\text{MLE}} = {{\sum_{i=1}^n I_{\{x_1^i=x_2^i\}}} \over {n}}.
\end{equation}

We can get $\hat{\theta}_{\text{MLE}}$ from maximum likelihood estimation for an MRF. 
Although the exact likelihood function is not available, a few algorithms \citep{geyer91,zhu02,hinton02,tieleman08,tieleman09,asuncion10b} make use of the concavity of the MRF's log likelihood function and find the MLE via gradient ascent with both satisfactory empirical performance and convergence properties \citep{younes88,yuille04,carreira05,salakhutdinov09,sutskever10}.
To find MLE of $\theta$, we use the contrastive divergence algorithm \citep{hinton02} or its variations \citep{tieleman08,tieleman09,asuncion10b} which are essentially gradient ascent algorithms. 
Note that the gradient ascent algorithms estimate the MLE from the joint likelihood function. 
Theoretically, the components of the MLE from the joint likelihood function may not necessarily be the same as the MLEs from the marginal likelihood functions, but the difference is expected to be reasonably  small, especially when the joint likelihood function is not highly skewed or multi-modal.

With $\hat{\lambda}_{\text{MLE}}$ and $\hat{\theta}_{\text{MLE}}$ available, $\eta_0$ could be calculated by Formula (\ref{eq4}).
The marginal likelihood function of $\theta$ can be approximated by (\ref{eq:specialbeta}) with $\alpha_0 = \sum_{i=1}^n I_{\{x_1^i=x_2^i\}}$. 
As one may tell, all the calculation does not depend on the specific structure of the MRF. 
However, our approximation is valid because how one parameter interacts with the remaining parameters in the likelihood function is properly captured.  
Computationally, the approximation of the marginal likelihood functions has no further computational cost other than performing contrastive divergence to find $\hat{\theta}_{\text{MLE}}$. 

Note that this is how we handle $\theta$ in the pairwise potential functions on the edges. 
In the situation where $\theta$ parameterizes a node potential, the methodology is also applicable. 
In the situation where there are more than two possible outcomes, we need to generalize both Dirichlet distribution and multinomial distribution in a similar fashion.

\subsection{Joint Likelihood Reconstruction From Marginal Likelihood}\label{sec:copula}

The marginal likelihood functions provide building blocks for the reconstruction of the joint likelihood function. 
Sklar's Theorem guarantees that any multivariate joint distribution be written in terms of univariate marginal distribution functions and a copula which describes the dependence structure between the variables \citep{sklar59}. 
With the marginal likelihood functions available, what we need is a proper copula which specifies the dependence structure between individual parameters \citep{nelsen06}.

It is rather challenging to capture the accurate copula function given the MRF structure, and the difficulty is expected to grow fast as the dimension of parameter space increases. 
While we prefer a copula function that is as close to the ground truth as possible, it could be impractical to pursue such an overly complicated approximation. 
We aim at a balance between accuracy and feasibility, and consider the conventional copula functions with fewer parameters.

Two types of copula functions are investigated in this work. 
One is the independent copula, and the other one is the exchangeable Gaussian copula. 
The independent copula assumes that the individual marginal likelihood functions in terms of the parameters are independent. 
The joint likelihood function is simply the product of these marginal likelihood functions. 
Namely, if $L(\theta_j|\mathbb{X})$ is the marginal likelihood function for $\theta_j$ $(1 \leq j \leq p)$, then the joint likelihood function is given as follows

\begin{equation} \label{eq6}
L({\boldsymbol \theta}|\mathbb{X}) = \prod_{j=1}^{p} L(\theta_j|\mathbb{X}).
\end{equation}

With an exchangeable Gaussian copula, a correlation matrix $R$ needs to be pre-specified. 
$R$ is a $p {\times} p$ positive definite matrix, whose off-diagonal entries are all equal to a certain value $\rho$ \citep{wu07}. 
The copula density in terms of transformed variables is shown below,

\begin{equation} \label{eq7}
c_{R}^{Gauss}({\boldsymbol u({\boldsymbol \theta})}) = {{1}\over{\sqrt{\det R}}} \exp \{-{{1}\over{2}} {\boldsymbol u({\boldsymbol \theta})}^T (R^{-1}-I){\boldsymbol u({\boldsymbol \theta})} \},
\end{equation}
where ${\boldsymbol u({\boldsymbol \theta})} = c(u_1,\ldots,u_p)$ and $u_j = \Phi^{-1} (F_j(\theta_j|\mathbb{X}))$ for $1 \leq j \leq p$.
Here $\Phi^{-1}()$ is the inverse of cumulative distribution function for the standard normal distribution, and $F_j()$ is the cumulative distribution function from standardizing the marginal likelihood function of $\theta_j$, such as the particular likelihood from the modified scenario of coin tossing. 
When $\rho=0$, $R$ becomes an identity matrix and this Gaussian copula reduces to the independent copula. 
Given the exchangeable Gaussian copula as well as the marginal likelihood functions, the joint likelihood is specified as
\begin{equation} \label{eq8}
L({\boldsymbol \theta}|\mathbb{X}) = c_{R}^{Gauss}({\boldsymbol u({\boldsymbol \theta})}) \prod_{j=1}^{p} L(\theta_j|\mathbb{X}).
\end{equation}

The simulation results in the next section suggest that the empirical performance of our MLE-induced likelihood is insensitive to the choice of $\rho$ in the exchangeable copula function.

\section{Simulations}\label{sec:simulation}

In this section, we empirically evaluate our MLE-induced likelihood with two different types of approximation --- pseudolikelihood \citep{besag75} and Laplace approximation \citep{welling06}.
Since the likelihood function is heavily used in Bayesian analysis, we evaluate them in MRF Bayesian learning tasks.
Specifically, we compare them in a widely used Markov chain Monte Carlo algorithm --- the standard Metropolis-Hastings (MH) algorithm \citep{metropolis53,hastings70}.
Note that new types of sampling methods can be used for MRF Bayesian learning, such as Hamiltonian \citep{neal11}, Langevin \citep{murray04}, herding \citep{chen13} and tempered transitions \citep{salakhutdinov09}, which usually lead to higher efficiency and a better convergence rate.
However since the scope of this paper is to compare different types of approximation of the MRF likelihood function, we do not include these advanced sampling algorithms other than the standard Metropolis-Hastings algorithm.
In Subsection \ref{sec:backgroundBayesian}, we review the setup of the MRF Bayesian parameter learning task and the Metropolis-Hastings algorithm.
In Subsection \ref{sec:simulationSetup}, we provide more details about the models and the simulations.
In Subsection \ref{sec:simulationResults}, the empirical results from these different methods are provided and discussed.

\subsection{Bayesian Parameter Learning of MRFs} \label{sec:backgroundBayesian}

Suppose that an MRF on ${\bf X} {\in} \mathcal{X}^d$ ($\mathcal{X}$ is a discrete space) is parameterized by ${\boldsymbol \theta}$, and its probability mass function is $P({\bf X};{\boldsymbol \theta}) {=} {{\tilde{P}({\bf X};{\boldsymbol \theta})} / Z({\boldsymbol \theta})} $, where $\tilde{P}({\bf X};{\boldsymbol \theta})$ is some unnormalized probability measure, and $Z({\boldsymbol \theta})$ is the normalizing constant, a.k.a.~the partition function.
Given a prior of ${\boldsymbol \theta}$ and $n$ i.i.d.~observed data points $\mathbb{X}{=}\{ {\bf x}^1,...,{\bf x}^n\}$, Bayesian parameter estimation provides the posterior distribution of ${\boldsymbol \theta}$. This posterior distribution is essential in Bayesian inference. The expected value $E({\boldsymbol \theta}|\mathbb{X})$ (a.k.a.~Bayesian estimate or posterior mean) is the optimal estimate under many settings, and the standard deviation of the poster distribution further quantifies the variability of such optimal estimate.

Bayesian parameter estimation for general MRFs is known as doubly-intractable \citep{murray06}. With a prior $\pi({\boldsymbol \theta})$ and one data point ${\bf x}$, the posterior is $P({\boldsymbol \theta}|{\bf x}) \propto \pi({\boldsymbol \theta}) \tilde{P}({\bf x};{\boldsymbol \theta})/Z({\boldsymbol \theta})$. If we use the Metropolis-Hastings (MH) algorithm to generate posterior samples of ${\boldsymbol \theta}$, then in each MH step we have to calculate the MH ratio for the proposal from ${\boldsymbol \theta}$ to ${\boldsymbol \theta}^*$

\begin{equation}\label{formulamhratio}
\begin{split}
a({\boldsymbol \theta}^*|{\boldsymbol \theta}) & = {{\pi({\boldsymbol \theta}^*) P({\bf x};{\boldsymbol \theta}^*)Q({\boldsymbol \theta}|{\boldsymbol \theta}^*)} \over {\pi({\boldsymbol \theta}) P({\bf x};{\boldsymbol \theta})Q({\boldsymbol \theta}^*|{\boldsymbol \theta})}} \\
&={{\pi({\boldsymbol \theta}^*) \tilde{P}({\bf x};{\boldsymbol \theta}^*)Q({\boldsymbol \theta}|{\boldsymbol \theta}^*)Z({\boldsymbol \theta})} \over {\pi({\boldsymbol \theta}) \tilde{P}({\bf x};{\boldsymbol \theta})Q({\boldsymbol \theta}^*|{\boldsymbol \theta})Z({\boldsymbol \theta}^*)}},
\end{split}
\end{equation}

where $Q({\boldsymbol \theta}^*|{\boldsymbol \theta})$ is some proposal distribution from ${\boldsymbol \theta}$ to ${\boldsymbol \theta}^*$, and with probability $\min\{1,a({\boldsymbol \theta}^*|{\boldsymbol \theta})\}$ we accept the move from ${\boldsymbol \theta}$ to ${\boldsymbol \theta}^*$.
The MH algorithm converges to the correct posterior distribution if the Markov chain is ergodic and in detailed balance.
However, even the MH ratio itself is generally intractable because of the intractable likelihood function.

Primarily, we approximate $P({\bf x};{\boldsymbol \theta}^*)$ and $P({\bf x};{\boldsymbol \theta})$ in the MH ratio with our MLE-induced likelihood, pseudolikelihood \citep{besag75} and Laplace approximation \citep{welling06} and compare the performance of the Bayesian estimator when coupled with these three different types of approximation.
We are also aware of other sampling-based methods which make the calculation of the MH ratio feasible via estimating $Z({\boldsymbol \theta})/Z({\boldsymbol \theta}^*)$, including importance sampling \citep{meng96}, auxiliary variables \citep{moller06}, exchange algorithm \citep{murray06}, and persistent Markov chains \citep{chen12}. 
Although estimating the MH ratio is a broader goal, we still include importance sampling \citep{meng96}, auxiliary variables \citep{moller06}, exchange algorithm \citep{murray06}, and persistent Markov chains \citep{chen12} as secondary baselines in the simulations. 

\subsection{Simulation Setup}\label{sec:simulationSetup}

We investigate the empirical performance of these different Bayesian estimation methods with three models of different sizes: (i) a $4 {\times} 4$ grid MRF with heterogeneous parameters, (ii) a $6 {\times} 6$ grid MRF with heterogeneous parameters and (iii) an $8 {\times} 8$ grid MRF with heterogeneous parameters.

In total, we compare {\em ten} different Bayesian estimation methods, including {\em six likelihood-function-based methods} --- the exact likelihood calculation (Exact-L), our MLE-induced likelihood from an independent copula (MLE-L, $\rho{=}0$), our MLE-induced likelihood from a Gaussian copula with $\rho{=}0.05$ (MLE-L, $\rho{=}0.05$), our MLE-induced likelihood from a Gaussian copula with $\rho{=}0.1$ (MLE-L, $\rho{=}0.1$), pseudolikelihood approximation (Pseudo-L), Laplace approximation (Laplace-L), and {\em four sampling-based baseline methods} --- importance sampling with geometric $\alpha$ (IS-Geometric), auxiliary variable (AuxVar), exchange (Exch) and persistent Markov chain (PersistMC).
For the exact likelihood calculation, we do not enumerate and sum over all possible instantiations for $Z({\boldsymbol \theta})$, which has a complexity of $O(|\mathcal{X}|^d)$.
Instead, we use the efficient recursive algorithm \citep{reeves04} which also provides exact calculation of $Z({\boldsymbol \theta})$ and reduces the computation complexity to $O(|\mathcal{X}|^{l+1})$, $l=\min\{\# rows, \# columns\}$ for grid-structured MRFs.
The reason we choose the geometric function for $\alpha$ in the standard importance sampling algorithm is that it is parameter-free and yields accurate estimate of partition function ratios \citep{meng96}.
More algorithmic details of the four secondary baselines are provided in the supplementary material.

In the simulations, we first set the parameters, and then generate a number of data points under the parameters.
We then run the MH algorithm coupled with these different methods of calculating the MH ratio.
When we couple these different methods with the MH algorithm, we set the Markov chains in the same way for a fair comparison, including the initialization of the Markov chain, the number of MH steps and the proposal distribution $Q$.
For the four sampling-based methods (IS-Geometric, AuxVar, Exch and PersistMC), we use the same number of particles.

\subsection{Simulation Results}\label{sec:simulationResults}

In the first set of simulations, we use a $4 {\times} 4$ grid-structured MRF with a set of heterogeneous parameters, i.e. ${\boldsymbol \theta}=\{\theta_1,...,\theta_p\}$, $p=24$.
The MRF is binary and pairwise. The pairwise potential function on edge $j$ is
$\theta_j^{I_{\{X_u=X_v\}}}(1-\theta_j)^{I_{\{X_u \not=X_v\}}}$ where nodes $X_u$ and $X_v$ are connected by edge $j$ in the graph.
We set $\theta_j$ uniformly distributed on the interval $(0.5, 0.8)$ for $j=1,\ldots,p$.
We then generate $n$ data points $(n{=}100, 200,..., 1{,}000)$.
Eventually, we apply the MH algorithm with ten different methods of calculating the MH ratio, namely Exact-L, MLE-L$(\rho{=}0)$, MLE-L$(\rho{=}0.05)$, MLE-L$(\rho{=}0.1)$, Pseudo-L, Laplace-L, IS-Geometric, AuxVar, Exch and PersistMC. 
We use a uniform distribution for the prior $\pi({\boldsymbol \theta})$.
The $Q({\boldsymbol \theta}^*|{\boldsymbol \theta})$ in (\ref{formulamhratio}) is a $p$-variate Gaussian density function with mean ${\boldsymbol \theta}$ and covariance matrix $\sigma_Q^2 I_{p}$. The $\sigma_Q^2$ is set to be $0.001$ in the simulations. 
The total number of MH steps for all methods is set to be $1{,}000{,}000$.
For the four sampling-based methods (IS-Geometric, AuxVar, Exch and PersistMC), we set the number of particles to be $1{,}000$.
When we have to generate particles from $P({\bf X};{\boldsymbol \theta})$ for a given ${\boldsymbol \theta}$, we randomly instantiate the particles and advance the particles for $1{,}000$ steps under ${\boldsymbol \theta}$.
We observe that coalescence \citep{propp96} always happens within the first $1{,}000$ steps, so the particles can be regarded as exact samples. As one may tell, we set the numbers in a conservative manner, minimizing the impact caused by simulation settings and placing our focus on the genuine performances of methods in comparison.

We use the posterior samples generated from exact likelihood as the benchmark, and compare the posterior samples generated from the other methods by two measures. 
The first is the posterior mean estimated from these methods.
The second is the posterior standard deviation estimated from these methods. 
We keep records of the run time for the methods in the simulations. 
All of the simulations have been replicated $50$ times and the averaged results are provided.

\begin{table}[t]
\caption{The run time (in milliseconds) of Exact-L, Pseudo-L, Laplace-L, our MLE-L$(\rho{=}0)$, our MLE-L$(\rho{=}0.05)$, our MLE-L$(\rho{=}0.1)$, IS-Geometric, AuxVar, Exch and PersistMC in Simulation 1. Algorithms with * were not finished and the run time was estimated from the MH steps actually finished.}
\label{table:runningTime:sim1}
\begin{center}
\begin{small}
\begin{sc}
\begin{tabular}{crrr}
\hline

  & $n=100$ & $n=500$ & $n=1{,}000$ \\
\hline

{Exact-L}&{1.19E+08}&{1.24E+08}&{1.34E+08}\\
{Pseudo-L}&{4.63E+05}&{2.35E+06}&{4.57E+06}\\
{Laplace-L}&{8.03E+03}&{8.07E+03}&{8.08E+03}\\
{MLE-L$(\rho{=}0)$}&{1.24E+04}&{1.25E+04}&{1.25E+04}\\
{MLE-L$(\rho{=}0.05)$}&{6.90E+04}&{6.92E+04}&{6.92E+04}\\
{MLE-L$(\rho{=}0.1)$}&{6.90E+04}&{6.91E+04}&{6.92E+04}\\
{IS-Geometric*}&{9.82E+09}&{9.85E+09}&{9.95E+09}\\
{AuxVar*}&{9.91E+09}&{1.00E+10}&{1.01E+10}\\
{Exch*}&{5.05E+09}&{5.06E+09}&{5.06E+09}\\
{PersistMC}&{2.83E+06}&{3.33E+06}&{4.00E+06}\\

\hline
\end{tabular}
\end{sc}
\end{small}
\end{center}
\end{table}

\begin{figure*}[t]
\begin{center}
\includegraphics[trim=0cm 0cm 0cm 0cm, clip=true, angle=0, scale=0.43]{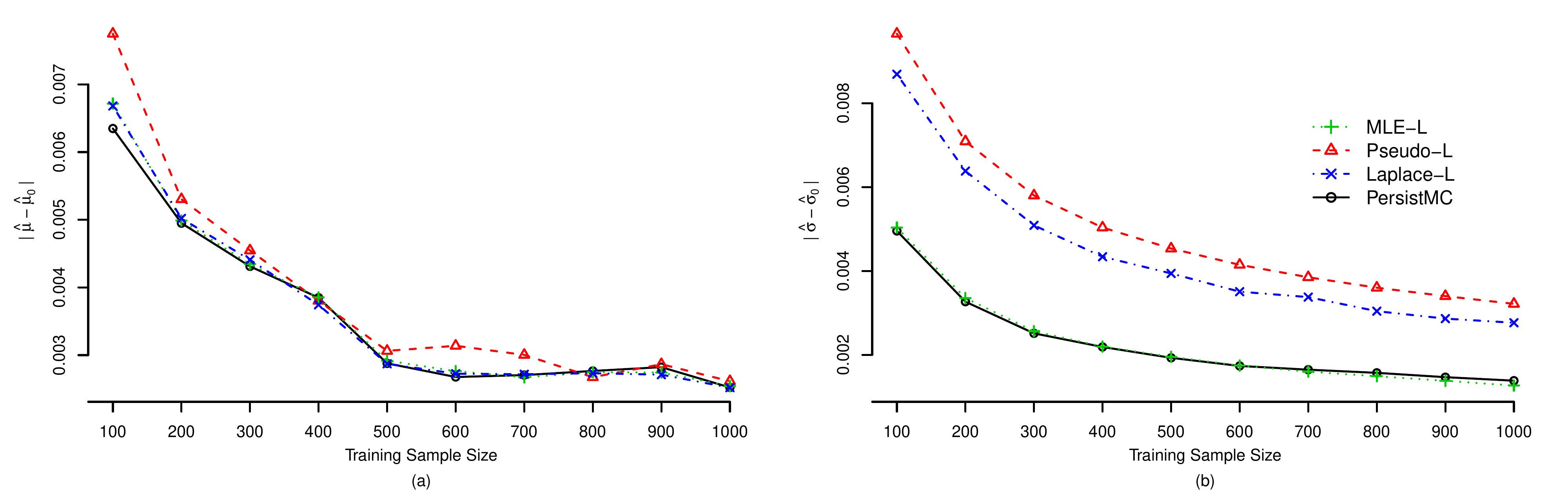}
\end{center}
\caption{Performance of MLE-L$(\rho{=}0)$, Pseudo-L, Laplace-L and PersistMC in Simulation 1 in terms of (a) the posterior mean estimate $\hat{\mu}$ and (b) the standard deviation estimate $\hat{\sigma}$, compared with the estimates ($\hat{\mu}_0$ and $\hat{\sigma}_0$) from exact likelihood calculation.}
\label{fig:sim1}
\end{figure*}

{\bf Run Time}: The run times of these methods in the first set of simulations are listed in Table \ref{table:runningTime:sim1}.
All the experiments are carried out on one 3GHz Intel Xeon CPU.
Three baselines, IS-Geometric, AuxVar and Exch, did not finish due to their long run time.
We estimate their run time based on how many MH steps they actually finished within the given time.
It is expected that they can finish the $1{,}000{,}000$ MH steps in thousands of hours.
Persistent Markov chain is the only sampling-based method that finished in a reasonable time (around an hour).
The run time of pseudolikelihood is roughly on the same level, but it is more sensitive to the number of observed data points.
The exact likelihood calculation is tolerable because we use the efficient recursive algorithm \citep{reeves04}.
Laplace approximation and our MLE-induced likelihood take the least time, and the run time is insensitive to the sample size.
For the different choices of the copula function, we observe a number of folds increase in run time if a Gaussian copula with a nonzero $\rho$ is selected over the independent copula.

{\bf Empirical Performance}: Three baselines (IS-Geometric, AuxVar and Exch) did not finish due to their long run time, and therefore their empirical performance cannot be evaluated.
The empirical performance of pseudolikelihood, Laplace approximation, persistent Markov chain and our MLE-L$(\rho{=}0)$ are plotted in Figure \ref{fig:sim1}.
The performance of our MLE-L is insensitive to the choice of the copula function.
Therefore, we do not plot MLE-L$(\rho{=}0.05)$ and MLE-L$(\rho{=}0.1)$ in the figure.
For the posterior mean estimated from the posterior samples, it is observed that all of the four methods yield satisfactory results.
As the number of data points increases, the posterior mean estimate from these Bayesian estimators (pseudolikelihood, Laplace approximation, persistent Markov chain and our MLE-L) gets closer to that from exact likelihood estimation.
Persistent Markov chain, Laplace approximation and our MLE-L yield comparable results on this small $4 {\times} 4$ grid MRF example.
Pseudolikelihood works slightly worse when there are less data points \footnote{The reason is that pseudolikelihood estimators are biased with finite samples, but are asymptotically consistent \citep{comets92}.}.
For the estimation of the posterior standard deviation, all the methods provide similar estimates as exact likelihood and the difference decreases as the number of data points increases.
Our MLE-L and persistent Markov chain perform similarly, and yield smaller estimation errors for posterior standard deviation.
Laplace approximation yields a larger estimation error for posterior standard deviation, and pseudolikelihood yields the largest.

{\bf Simulations In Higher-dimensional Parameter Space}: We replicate the experiments on the $6 {\times} 6$ grid-structured MRF with a set of heterogeneous parameters $(p=60)$ and the $8 {\times} 8$ grid-structured MRF with a set of heterogeneous parameters $(p=112)$.
We exclude IS-Geometric, AuxVar and Exch due to their long run time.
Exact-L did not finish either, and we lose the benchmark to compare the empirical performance.
Therefore, we compare the Bayesian estimates from pseudolikelihood, Laplace approximation, persistent Markov chain, our MLE-L$(\rho{=}0)$, MLE-L$(\rho{=}0.05)$, and MLE-L$(\rho{=}0.1)$ with the ground truth of the parameters.
All the other configurations in the simulations remain the same.

The run times of the six methods (pseudolikelihood, Laplace approximation, persistent Markov chain, our MLE-L$(\rho{=}0)$, MLE-L$(\rho{=}0.05)$, and MLE-L$(\rho{=}0.1)$) on the $6 {\times} 6$ grid MRF and the $8 {\times} 8$ grid MRF are listed in Table \ref{table:runningTime:sim2}.
Similar to previous simulations on the $4 {\times} 4$ grid MRF, Laplace-L and our MLE-L run the fastest, and the run time is insensitive to the number of data points.
For the different choices of the copula function in our MLR-L, there is a number of folds increase in run time if a Gaussian copula with a nonzero $\rho$ is chosen over the independent copula.
The run times of both Pseudo-L and PersistMC are sensitive to the number of data points, and get almost unbearable as both the number of data points increases and the number of parameters increases.

\begin{figure*}[t]
\begin{center}
\includegraphics[trim=0cm 0cm 0cm 0cm, clip=true, angle=0, scale=0.4]{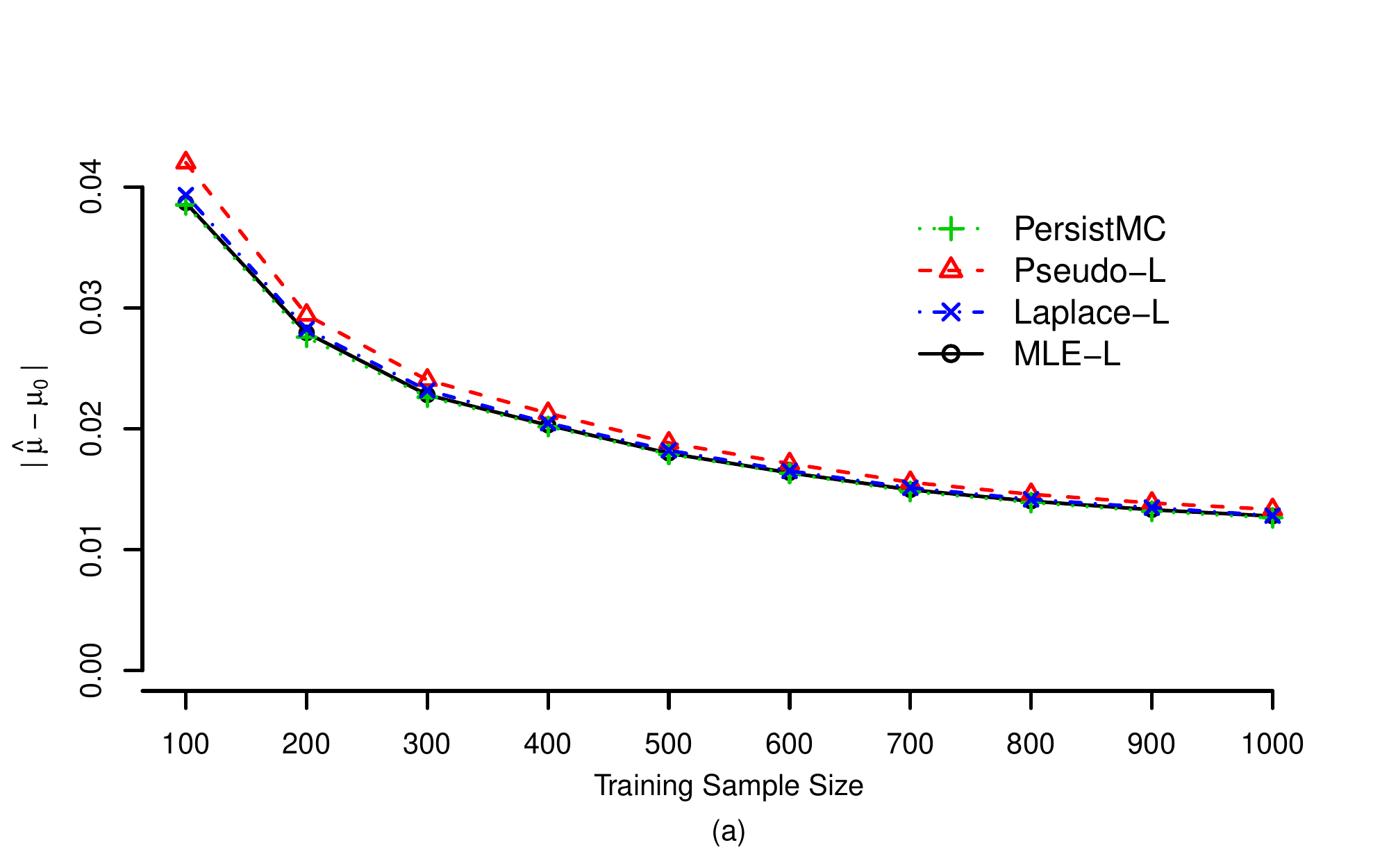} 
\includegraphics[trim=0cm 0cm 0cm 0cm, clip=true, angle=0, scale=0.4]{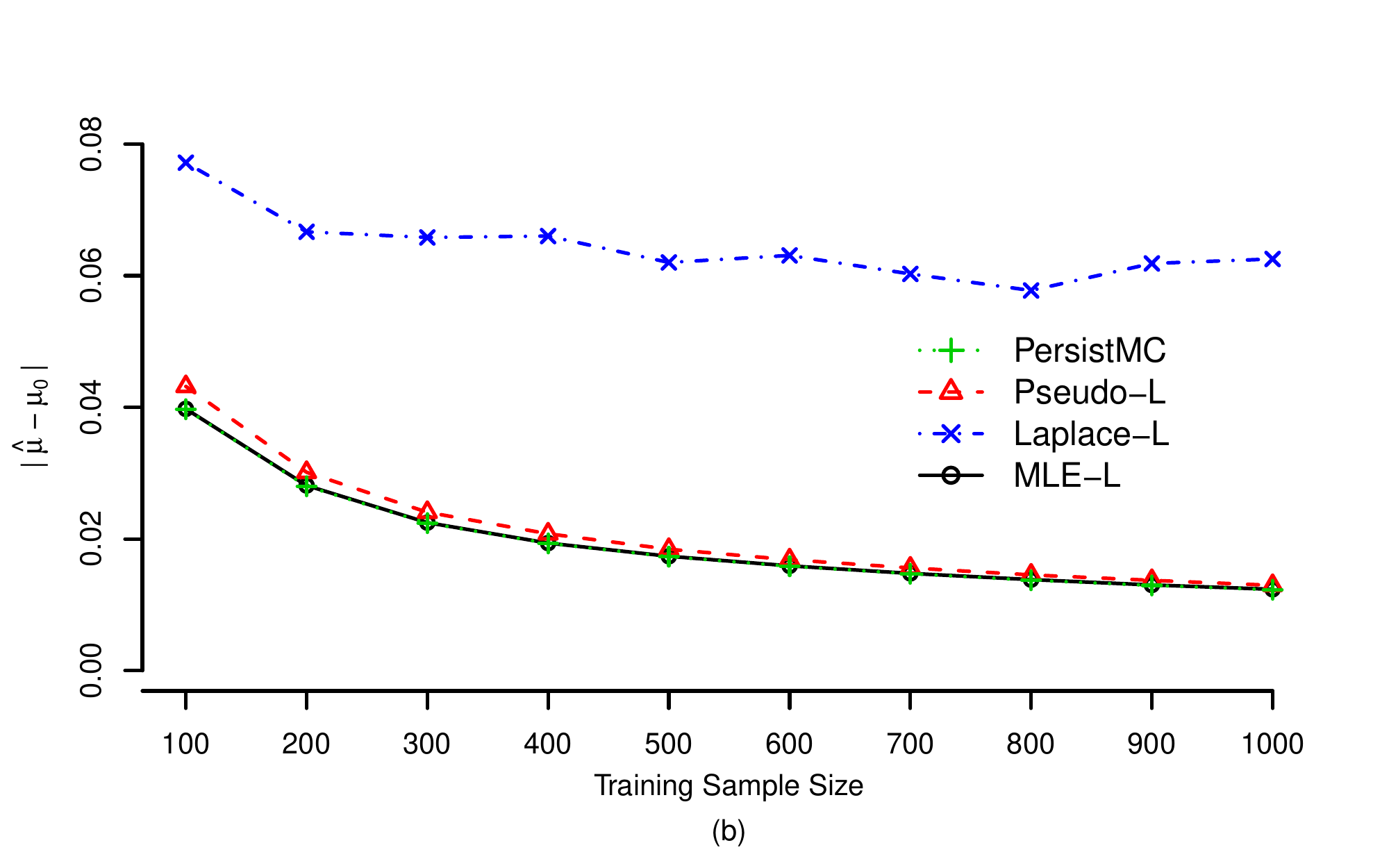} 
\end{center}
\caption{The posterior mean estimate of MLE-L$(\rho{=}0)$, Pseudo-L, Laplace-L and PersistMC in (a) $6 {\times} 6$ grid-structured MRF and (b) $8 {\times} 8$ grid-structured MRF.}
\label{fig:sim2}
\end{figure*}

The performance of the six methods (pseudolikelihood, Laplace approximation, persistent Markov chain, our MLE-L$(\rho{=}0)$, MLE-L$(\rho{=}0.05)$, and MLE-L$(\rho{=}0.1)$) is evaluated by the difference of yielded Bayesian estimates and the ground truth of the parameters.
The numerical results are plotted in Figure \ref{fig:sim2}.
The performance of pseudolikelihood, persistent Markov chain and our MLE-L$(\rho{=}0)$ on the larger MRFs is similar to that from the $4 {\times} 4$ grid MRF.
The performance of our MLE-L is also insensitive to the choice of the copula function.
Therefore, we do not plot MLE-L$(\rho{=}0.05)$ and MLE-L$(\rho{=}0.1)$ in the figure.
However, the performance of Laplace approximation quickly deteriorates as the dimension of parameter space grows, which suggests Laplace approximation is less favorable than others in large-scale MRFs.

\begin{table}[t]
\caption{The run time (in milliseconds) of PersistMC, Pseudo-L, Laplace, our MLE-L$(\rho{=}0)$, MLE-L$(\rho{=}0.05)$, and MLE-L$(\rho{=}0.1)$ in the simulations on the $6 {\times} 6$ grid MRF and the $8 {\times} 8$ grid MRF.}
\label{table:runningTime:sim2}
\begin{center}
\begin{small}
\begin{sc}
\begin{tabular}{crrr}
\hline

  & $n=100$ & $n=500$ & $n=1{,}000$ \\
\hline

{$6 {\times} 6$ grid}&{}&{}&{}\\
{PersistMC}&{7.74E+06}&{9.67E+06}&{1.12E+07}\\
{Pseudo-L}&{4.81E+05}&{2.65E+06}&{5.00E+06}\\
{Laplace-L}&{2.41E+04}&{2.42E+04}&{2.42E+04}\\
{MLE-L$(\rho{=}0)$}&{2.70E+04}&{2.72E+04}&{2.73E+04}\\
{MLE-L$(\rho{=}0.05)$}&{1.62E+05}&{1.63E+05}&{1.65E+05}\\
{MLE-L$(\rho{=}0.1)$}&{1.62E+05}&{1.62E+05}&{1.64E+05}\\

\hline

{$8 {\times} 8$ grid}&{}&{}&{}\\
{PersistMC}&{1.45E+07}&{1.78E+07}&{1.98E+07}\\
{Pseudo-L}&{2.25E+06}&{1.02E+07}&{1.77E+07}\\
{Laplace-L}&{9.90E+04}&{9.92E+04}&{9.93E+04}\\
{MLE-L$(\rho{=}0)$}&{1.22E+05}&{1.24E+05}&{1.27E+05}\\
{MLE-L$(\rho{=}0.05)$}&{1.05E+06}&{1.06E+06}&{1.07E+06}\\
{MLE-L$(\rho{=}0.1)$}&{1.04E+06}&{1.06E+06}&{1.07E+06}\\

\hline
\end{tabular}
\end{sc}
\end{small}
\end{center}
\end{table}

\section{Discussion and Conclusion}\label{sec:conclusion}

The primary contribution of our paper is proposing a two-step method of approximating the MRF likelihood function.
The proposed MLE-induced likelihood yields satisfactory numerical precision during the approximation.
Especially as the size of the MRF increases, both the numerical performance and the computational cost of our approach remain consistently satisfactory, whereas Laplace approximation deteriorates and pseudolikelihood becomes computationally unbearable.
We also demonstrate that when our MLE-induced likelihood is used for MRF Bayesian parameter learning, its numerical performance is almost as good as the state-of-the-art sampling-based method --- persistent Markov chain \citep{chen12}, but the computational cost is significantly reduced.

From the methodology aspect, our MLE-induced likelihood is desirable for several reasons. 
We break down the approximation into two steps. 
In the first step, we only focus on the marginal likelihood functions and make sure that they are accurately approximated.
The particular likelihood function (derived from the modified scenario of coin tossing) properly captures how one parameter interacts with the remaining parameters (as a whole) in the likelihood function.
This is more effective than modeling how one parameter interacts with the remaining parameters individually (as in Laplace approximation).
During the calculation of the marginal likelihood functions, we also make use of the MLE of the parameters (e.g. from the contrastive divergence algorithm \citep{hinton02}), which makes sure that the modes of the marginal likelihood functions are accurate.
The second step of reconstruction is supported by Sklar's Theorem, which preserves the precision of the individual marginal likelihood functions and guarantees that the reconstructed joint likelihood is marginal-wise accurate. 

In addition to the satisfactory empirical performance and the methodology innovation, the proposed MLE-induced likelihood also has some connection with some known distributions.
Especially, the marginal likelihood function we derive in Formula (\ref{eq:specialbeta}) is related to the density function of a Generalized Beta Distribution (GBD) \citep{mcdonald95}.
If the power in the denominator of (\ref{eq:specialbeta}) is $n+2$ instead of $n$, function (\ref{eq:specialbeta}) can be standardized to be a density function of a GBD by dividing a constant.
In general, $n$ is relatively large, and $n+2$ and $n$ are quite close, and (\ref{eq:specialbeta}) is proportional to a GBD density function approximately.

Last but not least, our proposed MLE-induced likelihood nicely extends the recent advances within the field of Markov random field learning. 
Although the typical MLE procedures require that we first specify the log likelihood function, and then find the MLE of the parameters, it does not mean that we cannot find the MLE for MRFs. 
Indeed, a few recent algorithms \citep{geyer91,zhu02,hinton02,tieleman08,tieleman09,asuncion10b} make use of the concavity of the MRF's log likelihood function and find the MLE via gradient ascent with both satisfactory empirical performance \citep{liu12b,liu2013structure,liu2013bayesian,liu2014learning,liu2014multiple,liu2016multiple} and convergence properties \citep{younes88,yuille04,carreira05,salakhutdinov09,sutskever10}. 
With the availability of MLE of the parameters, we further demonstrate that we are able to recover the intractable likelihood function to some precision from the MLE of the parameters.

\bibliographystyle{plainnat}
\bibliography{references}

\section{Supplementary: Algorithmic Details of the Four Secondary Baselines}

The real hurdle in Bayesian parameter estimation for general MRFs is the intractable MH ratio 

\begin{equation}\label{formulamhratio}
\begin{split}
a({\boldsymbol \theta}^*|{\boldsymbol \theta}) & = {{\pi({\boldsymbol \theta}^*) P({\bf x};{\boldsymbol \theta}^*)Q({\boldsymbol \theta}|{\boldsymbol \theta}^*)} \over {\pi({\boldsymbol \theta}) P({\bf x};{\boldsymbol \theta})Q({\boldsymbol \theta}^*|{\boldsymbol \theta})}} \\
&={{\pi({\boldsymbol \theta}^*) \tilde{P}({\bf x};{\boldsymbol \theta}^*)Q({\boldsymbol \theta}|{\boldsymbol \theta}^*)Z({\boldsymbol \theta})} \over {\pi({\boldsymbol \theta}) \tilde{P}({\bf x};{\boldsymbol \theta})Q({\boldsymbol \theta}^*|{\boldsymbol \theta})Z({\boldsymbol \theta}^*)}}.   
\end{split} 
\end{equation}

We can approximate $P({\bf x};{\boldsymbol \theta}^*)$ and $P({\bf x};{\boldsymbol \theta})$ in the MH ratio with our MLE-induced likelihood, pseudolikelihood \citep{besag75} and Laplace approximation \citep{welling06}.
Other than approximating the likelihood functions, we can also use other sampling-based methods which make the calculation of the MH ratio feasible via estimating $Z({\boldsymbol \theta})/Z({\boldsymbol \theta}^*)$, including importance sampling \citep{meng96}, auxiliary variables \citep{moller06}, exchange algorithm \citep{murray06} and persistent Markov chains \citep{chen12}.
Here we provide the algorithmic details of the four secondary baselines.

\subsection{Importance Sampling}

We can use importance sampling to estimate $r=Z({\boldsymbol \theta})/Z({\boldsymbol \theta}^*)$ \citep{meng96} by

\vspace*{-0.05in}
\begin{equation}\label{r_is_alpha}
\hat{r}_{IS} = { {  {s_2^{-1}} \sum_{t=1}^{s_2} \tilde{P}({\bf x}_2^{(t)};{\boldsymbol \theta}) \alpha({\bf x}_2^{(t)}) } \over { {s_1^{-1}} \sum_{t=1}^{s_1} \tilde{P}({\bf x}_1^{(t)};{\boldsymbol \theta}^*) \alpha({\bf x}_1^{(t)})  } },
\end{equation}
\vspace*{-0.05in}

where ${\bf x}_1^{(1)}$, ..., ${\bf x}_1^{(s_1)}$ are sampled from $P({\bf X};{\boldsymbol \theta})$ and ${\bf x}_2^{(1)}$, ..., ${\bf x}_2^{(s_2)}$ are sampled from $P({\bf X};{\boldsymbol \theta}^*)$, and $\alpha({\bf X})$ is an arbitrary function defined on the same support as $\tilde{P}$. 
Theoretically, $\hat{r}_{IS}$ is a consistent estimator of $Z({\boldsymbol \theta})/Z({\boldsymbol \theta}^*)$ as long as the sample averages in (\ref{r_is_alpha}) converge to their corresponding population averages, which is satisfied by Markov chain Monte Carlo under regular conditions. 
However, the optimal choice of $\alpha$ depends on the ground truth of $r$, and \citep{meng96} provides several options for $\alpha$, such as a geometric function $\alpha({\bf X})=(\tilde{P}({\bf X};{\boldsymbol \theta})\tilde{P}({\bf X};{\boldsymbol \theta}^*))^{-1/2}$.

\subsection{Auxiliary Variables}

The second method is to introduce auxiliary variables and cancel $Z({\boldsymbol \theta})/Z({\boldsymbol \theta}^*)$ in (\ref{formulamhratio}). 
\citep{moller06} introduces one auxiliary variable ${\bf Y}$ on the same space as ${\bf X}$, and the state variable is extended to $({\boldsymbol \theta},{\bf Y})$. 
They set the new proposal distribution for the extended state $Q({\boldsymbol \theta},{\bf Y}|{\boldsymbol \theta}^*{,}{\bf Y}^*) {=} Q({\boldsymbol \theta}|{\boldsymbol \theta}^*)\tilde{P}({\bf Y};{\boldsymbol \theta})/Z({\boldsymbol \theta})$ to cancel $Z({\boldsymbol \theta})/Z({\boldsymbol \theta}^*)$ in (\ref{formulamhratio}). 
Therefore by ignoring ${\bf Y}$, we can generate the posterior samples of ${\boldsymbol \theta}$ via Metropolis-Hastings. 
Technically, this auxiliary variable approach requires perfect sampling \citep{propp96}, but \citep{moller06} pointed out that other simpler Markov chain methods also work with the proviso that they converge adequately to the equilibrium distribution. 
\citep{murray06} extended the single auxiliary variable method to multiple auxiliary variables for improved efficiency, as well as pointed out that the single auxiliary variable method can be simplified as a single-variable exchange algorithm. 
Both the single auxiliary variable algorithm and the single-variable exchange algorithm can be interpreted as importance sampling \cite{moller06,murray06}. 
In the auxiliary variable algorithm, $r=Z({\boldsymbol \theta})/Z({\boldsymbol \theta}^*)$ is estimated by

\vspace*{-0.05in}
\begin{equation}\label{r_auxvar}
\hat{r}_{aux} = { {   {s_1^{-1}}} \sum_{t=1}^{s_1} { {\tilde{P}({\bf x}_1^{(t)};\hat{\boldsymbol \theta})} \over {\tilde{P}({\bf x}_1^{(t)};{\boldsymbol \theta})} } \over {   {s_2^{-1}}} \sum_{t=1}^{s_2} {{\tilde{P}({\bf x}_2^{(t)};\hat{\boldsymbol \theta})} \over {\tilde{P}({\bf x}_2^{(t)};{\boldsymbol \theta}^*)} } },
\end{equation}
\vspace*{-0.05in}

where ${\bf x}_1^{(1)}$, ..., ${\bf x}_1^{(s_1)}$ are sampled from $P({\bf X};{\boldsymbol \theta})$ and ${\bf x}_2^{(1)}$, ..., ${\bf x}_2^{(s_2)}$ are sampled from $P({\bf X};{\boldsymbol \theta}^*)$, and $\hat{\boldsymbol \theta}$ is some estimate of ${\boldsymbol \theta}$. 

\subsection{Exchange Algorithm}

In the single-variable exchange algorithm, $r=Z({\boldsymbol \theta})/Z({\boldsymbol \theta}^*)$ is estimated by

\vspace*{-0.05in}
\begin{equation}\label{r_exch}
\hat{r}_{exch} = {{s^{-1}}} \sum_{t=1}^{s} {{\tilde{P}({\bf x}^{(t)};{\boldsymbol \theta})}\over{\tilde{P}({\bf x}^{(t)};{\boldsymbol \theta}^*)}}, 
\end{equation}
\vspace*{-0.05in}

where ${\bf x}^{(1)}$, ..., ${\bf x}^{(s)}$ are sampled from $P({\bf X};{\boldsymbol \theta}^*)$. 

\subsection{Persistent Markov Chains}\label{algorithm2}

Importance sampling, the auxiliary variable method and the exchange algorithm are computationally intensive and do not scale well to large models or high dimensional parameter space. It is because that in each MH step they require generating samples from $P({\bf X};{\boldsymbol \theta})$ for a given ${\boldsymbol \theta}$ via the computationally expensive perfect sampling \citep{propp96} or standard Gibbs sampling with long runs. 

In the standard single-variable exchange algorithm, $s$ samples need to be generated from $P({\bf X};{\boldsymbol \theta}^*)$ in each MH step when we propose ${\boldsymbol \theta}^*$, and the MH ratio is calculated as (\ref{r_exch}). The motivation of the persistent Markov chains algorithm \citep{chen12} is that although the proposed ${\boldsymbol \theta}^*$ is different from ${\boldsymbol \theta}$, ${\boldsymbol \theta}^*$ is usually not far away from ${\boldsymbol \theta}$ because MH algorithms usually require proposing small changes so as to maintain a high acceptance rate. Therefore the particles from ${\boldsymbol \theta}^*$ should be quite similar to the particles from ${\boldsymbol \theta}$. Therefore, we can reuse the particles generated from ${\boldsymbol \theta}$ in the previous step and further advance the particles for $k$ steps under the new parameter ${\boldsymbol \theta}^*$, where $k$ is a small number. Therefore, $r=Z({\boldsymbol \theta})/Z({\boldsymbol \theta}^*)$ in the $l$-th MH step can be estimated by

\vspace*{-0.05in}
\begin{equation}\label{r_exch_PST}
\hat{r}_{persistMC} = {{s^{-1}}} \sum_{t=1}^{s} {{\tilde{P}({\bf x}^{(t)};{\boldsymbol \theta})}\over{\tilde{P}({\bf x}^{(t)};{\boldsymbol \theta}^*)}}, 
\end{equation}
\vspace*{-0.05in}

where ${\bf x}^{(1)}$, ..., ${\bf x}^{(s)}$ are the particles from the $(l{-}1)$-th MH step and get advanced for $k$ steps under ${\boldsymbol \theta}^*$. In our experiments, we set $k$ to be 1. This technique was firstly used in the persistent contrastive divergence algorithm \citep{tieleman08} which modified the standard contrastive divergence algorithm by reusing the particles persistently to save computation.

\end{document}